\documentclass[runningheads]{llncs}

\usepackage{multirow}
\usepackage{graphicx}
\usepackage{graphicx}
\usepackage{amsmath}
\usepackage{amssymb}
\usepackage{booktabs}
\usepackage{multirow}
\usepackage{xcolor}
\usepackage{makecell}
\usepackage{enumitem}
\usepackage{hyperref}

\usepackage{pifont}
\newcommand{\xmark}{\ding{55}}%
\newcommand{\cmark}{\ding{51}}%
\makeatletter
\newcommand{\printfnsymbol}[1]{%
  \textsuperscript{\@fnsymbol{#1}}%
}
\begin{document}
\title{Target and Task specific Source-Free Domain Adaptive Image Segmentation}

\author{Vibashan VS\hspace{0.1mm}\thanks{Equal contribution}, Jeya Maria Jose Valanarasu\hspace{0.1mm}\printfnsymbol{1}, and Vishal M. Patel } 
\institute{Johns Hopkins University}

%
\maketitle             
\begin{abstract} 
Solving the domain shift problem during inference is essential in medical imaging, as most deep-learning based solutions suffer from it. In practice, domain shifts are tackled by performing Unsupervised Domain Adaptation (UDA), where a model is adapted to an unlabelled target domain by leveraging the labelled source data. In medical scenarios, the data comes with huge privacy concerns making it difficult to apply standard UDA techniques. Hence, a closer clinical setting is Source-Free UDA (SFUDA), where we have access to source-trained model but not the source data during adaptation. Existing SFUDA methods rely on pseudo-label based self-training techniques to address the domain shift. However, these pseudo-labels often have high entropy due to domain shift and adapting the source model with noisy pseudo-labels leads to sub-optimal performance. To overcome this limitation, we propose a systematic two-stage approach for SFUDA comprising of target-specific adaptation followed by task-specific adaptation. In target-specific adaptation, we enhance the pseudo-label generation by minimizing high entropy regions using the proposed ensemble entropy minimization loss and a selective voting strategy. In task-specific adaptation, we exploit the enhanced pseudo-labels using a student-teacher framework to effectively learn segmentation on the target domain. We evaluate our proposed method on 2D fundus datasets and 3D MRI volumes across 7 different domain shifts where we perform better than existing UDA and SFUDA methods for medical image segmentation. Code is available at \href{https://github.com/Vibashan/tt-sfuda}{https://github.com/Vibashan/tt-sfuda}

\end{abstract}


\section{Introduction}

Medical image segmentation is an important task as it critical for image-guided surgery systems and computer aided diagnosis. In recent years, deep-learning based methods based on convolutional neural networks \cite{ronneberger2015u,milletari2016v,islam2018ischemic,valanarasu2020kiu} and transformers \cite{chen2021transunet,jose2021medical,hatamizadeh2022unetr} have been the leading solutions for medical image segmentation. One major concern with deep neural networks (DNNs) is that they are very much dependent on the dataset they are trained on. DNNs trained on a particular dataset do not perform well when tested on a different dataset even if it is of the same modality. Small changes in camera type, calibration properties, age and demographics of patients causes a distribution shift which results in a considerable drop in performance.  It is very common for a domain shift to exist during deployment of medical imaging solutions in real-world settings \cite{vashist2017point}. Hence  adapting a source model to a target data distribution is an important problem to solve for medical imaging applications like segmentation.

Recently, many works have explored unsupervised domain adaptation (UDA) for medical image segmentation \cite{javanmardi2018domain,panfilov2019improving,zhang2019unsupervised,yang2019unsupervised,perone2019unsupervised}. However, one major concern while applying UDA methods for medical imaging applications is that we assume we have the availability of labelled source domain data as well as unlabelled target domain data during adaptation. This assumption does not really hold true, especially for medical applications because medical data is sensitive and comes with a lot of privacy concerns. So, it is challenging to make use of the source data during adaptation time, considering the concerns in sharing medical data. To overcome this issue, Source-Free Unsupervised Domain adaptation (SFUDA) setting is explored where we assume we have access to only the source-trained model and the unlabelled target data for adaptation. This provides a practical clinical setting where we do not make use of the source data. SFUDA is relatively a new and challenging task and there have been very few works \cite{bateson2020source,liu2023memory,hong2022source} on this for medical image segmentation. In \cite{bateson2020source}, Bateson et al. proposes a SFUDA method where a label-free
entropy loss is reduced for target-domain data. Another recent study by Chen et al. in \cite{chen2021source} employed uncertainty estimation to generate denoised pseudo-labels for self-adaptation on target data. Reducing the entropy or denoising pseudo labels helps close the domain gap between the source and target data. Although these techniques aim to reduce the entropy of pseudo-labels to bridge the domain gap, our approach also emphasizes enhancing the quality of the generated pseudo-labels and effectively utilizing them to facilitate target domain segmentation learning.

In this work, we propose a novel systematic approach to tackle SFUDA, where we first enhance the quality of pseudo-labels and then effectively utilize them to improve the segmentation performance on the target domain. Specifically, we introduce a two-stage adaptation framework (\textbf{TT-SFUDA}): Stage I - Target-specific adaptation and Stage II - Task-specific adaptation. In target-specific adaptation, we  enhance the quality of pseudo-labels generated by the source-trained model. To achieve this, we propose an ensemble entropy minimization loss that minimizes the high entropy regions in the pseudo-labels generated by different data-augmented input images. Furthermore, we introduce a selective voting strategy that leverages the pseudo-labels generated by different data-augmented images to identify consistently high-entropy regions and generate high-quality pseudo-labels for better supervision. Overall, this leads to the generation of enhanced pseudo-labels, which are then exploited in Stage II of the adaptation process. In task-specific adaptation, we employ a student-teacher self-training framework using enhanced pseudo-labels, inspired by the semi-supervised method \cite{wang2021knowledge}. Here, we aim to enhance task-specific (i.e segmentation) information of the entropy-minimized model from Stage I. To achieve this, we apply strong and weak augmentation to the student and teacher model and ensure the predictions are consistent across the predictions. This helps the model learn target domain segmentation more effectively.  To validate our method, we conducted extensive experiments
on multiple domain shifts for both 2D and 3D medical imaging datasets and compare them with existing SFUDA and UDA adaptation methods.

In summary, this work makes the following contributions: \textbf{1)} We introduce a  systematic  SFUDA method focusing on both target and task-specific adaptation for image segmentation. \textbf{2)} We propose an efficient method for generating enhanced pseudo-labels using an ensemble entropy minimization loss and selective voting respectively. These improved pseudo-labels are then utilized to learn target domain segmentation more effectively. \textbf{3)} We conduct extensive experiments on both 2D and 3D data of different modalities. We show results on 7 different data-shifts for both 2D fundus images and 3D MRI volumes and achieve state-of-the-art performance compared to recent methods.   

\vspace{-1 em}
\section{Method}
\vspace{-1 em}
Our proposed TT-SFUDA method for medical image segmentation consists of two parts: Stage I - Target specific adaptation and Stage II - Task specific adaptation. First, we explain the base segmentation network architecture and formulate certain preliminaries and notations used in the proposed adaptation method.

\subsection{Preliminaries}
\setlength{\belowdisplayskip}{0pt} \setlength{\belowdisplayshortskip}{0pt}
\setlength{\abovedisplayskip}{0pt} \setlength{\abovedisplayshortskip}{0pt}
\noindent \textbf{Base Network Details:} We use a regular UNet \cite{ronneberger2015u} as the base segmentation network. We use a 5 level encoder-decoder framework. Each conv block in the encoder consists of a conv layer, ReLU activation and a max-pooling layer. Each conv block in the decoder consists of a conv layer, ReLU activation and an upsampling layer. For upsampling, we use bilinear interpolation. For our experiments on 3D volumes, we use a 3D UNet architecture \cite{cciccek20163d} with same setup replacing 2D conv layer with 3D conv layers, 2D max-pooling with 3D max-pooling and bilinear upsampling with trilinear upsampling. The segmentation loss used to train this network $\mathcal{L}_{seg}$ is as follows:
\begin{equation}
    \mathcal{L}_{seg} =  BCE(\hat{y}, y) + Dice(\hat{y}, y)
\end{equation}
\noindent \textbf{Notations:} We denote the input data as $x$ and labels as $y$. We denote the source domain data as $D_s = \{(x_s^n,y_s^n)\}_{n=1}^{N_s} $ and target domain data as  $D_t = \{(x_t^n,y_t^n)\}_{n=1}^{N_t}$ where $s$ represents source, $t$ represents target,  $N_s$ and $N_t$ denotes number of data-instances in source and target data respectively. We denote the source model of UNet as $\Theta_s$. The goal is to adapt $\Theta_s$ from  $D_s$ to  $D_t$ while using only the target data $x_t$. We assume we do not have access to $y_t$ which makes it unsupervised and also assume we do not have access to $D_s$ which makes it source-free. 

\begin{figure}[htbp]
	\centering
	\includegraphics[width=1.0\linewidth]{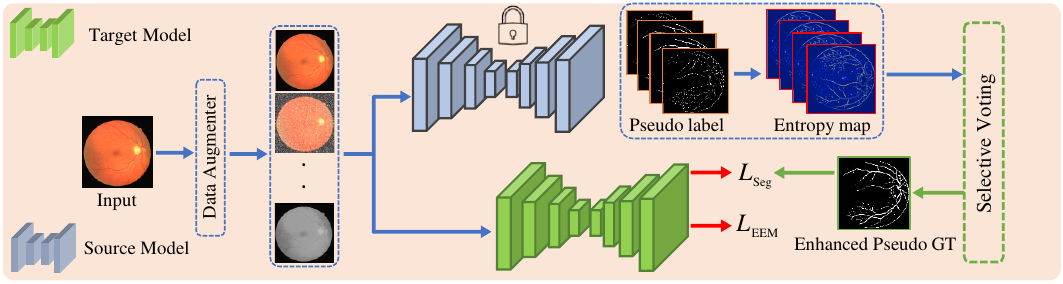}
	\caption{Overview of the proposed Target-specific adaptation method. }  
	\label{intro}
 \vspace{-0.5 em}
\end{figure}

\subsection{Target-specific Adaptation}
The main objective of the target-specific adaptation phase is to learn to enhance the pseudo-label generation using information from the source model. We define two instances of the model: source model ($\Theta_s$) and a target model ($\Theta_t^{stage1}$). Both these models are initialized with the source-trained model, but during adaptation, the source model ($\Theta_s$) is frozen while the target model ($\Theta_t^{stage1}$) is trained.  We achieve the objective of target-specific adaptation by utilizing the proposed Ensemble Entropy Minimization loss and selective voting for enhancing pseudo-labels.

\noindent \textbf{Ensemble Entropy Minimization: } For entropy minimization, we first generate an entropy map $H(\hat{y}_t^n)$ using output prediction map $\hat{y}_t^n$ by:
\begin{equation}
    H(\hat{y}_t^n) = - \sum p(\hat{y}_t^n) log (p(\hat{y}_t^n))
\end{equation}
To ensure that the model $\Theta_t^{stage1}$ generates better pseudo-labels, we aim to identify all high entropy regions in its predictions. However, using only one input data may not provide a complete picture of these regions. Therefore, we propose an ensemble entropy minimization approach to obtain a comprehensive understanding of the high entropy regions in the model. To achieve this, we first create a set of augmented data $\tilde{X}_t^n = \{ aug_i (x_t^n)\}_{i=1}^M$, where $M$ is the number of augmentations used. We then generate a corresponding set of pseudo-labels $\tilde{Y}_t^n$ for these augmented inputs using the source model ($\Theta_s$). Note that $\tilde{X}t^n$ and $\tilde{Y}t^n$ represent a set of data instances $\{x_t^n\}_{i=1}^M$ and $\{y_t^n\}_{i=1}^M$, respectively, depending on the number of augmentations used. Using the generated pseudo-labels $\tilde{Y}_t^n$, we compute entropy maps $H(\tilde{Y}_t^n)$ as:

\begin{equation}
    H(\tilde{Y}_t^n) = - \sum p(\tilde{Y}_t^n) \log (p(\tilde{Y}_t^n)).
    \label{eqn: ensem_ent}
\end{equation}
Now, we define the ensemble entropy minimization loss $\mathcal{L}_{EEM}$ as  follows:
\begin{equation}
   \mathcal{L}_{EEM} = \frac{1}{HW} \sum_{t=1}^{HW} ( H(\hat{y}_t^n) + \frac{1}{M} \sum_{j=1}^{M} H_j(\tilde{Y}_t^n) ).
\end{equation}
where $H$ and $W$ represent the height and width of the input image, respectively. Minimizing this loss helps suppress high entropy regions across different augmentations, leading to effective generation of pseudo-labels by the target model.

\begin{figure}[t]
	\centering
	\includegraphics[width=1.0\linewidth]{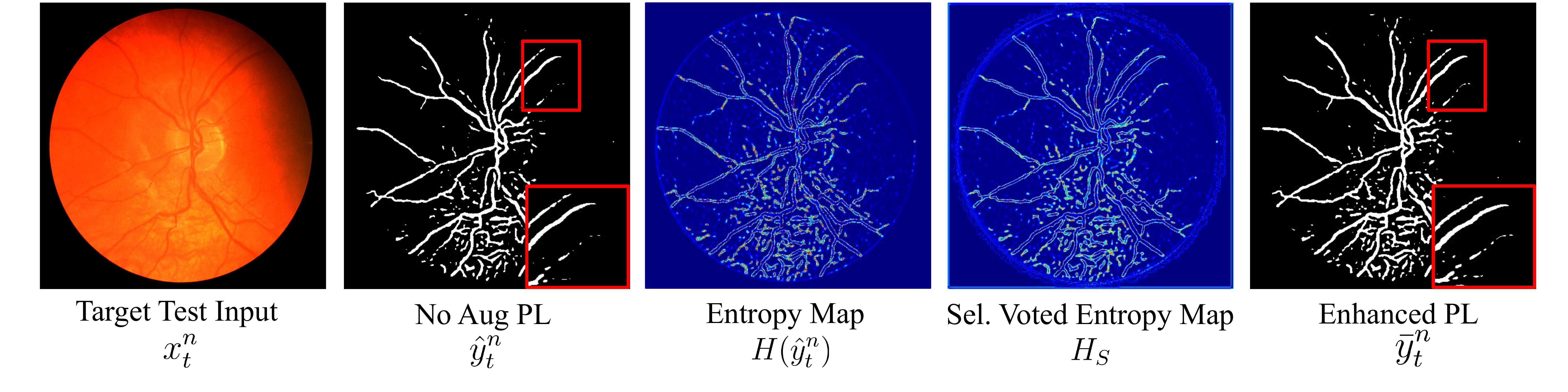}
	
	\vskip-10pt\caption{Steps involved in obtaining the enhanced pseudo label $\mathbf{\bar{y}_t^n}$. }  
	\label{stage1}
 \vspace{-0.5 em}
\end{figure}

\noindent \textbf{Selective Voting: } In addition to ensemble entropy minimization, we propose a selective voting strategy to further improve the quality of pseudo-labels. Specifically, this strategy aims to reduce false negative predictions in pseudo-label $\tilde{y}_t^n$ by utilizing the entropy maps $H(\hat{y}_t^n)$ obtained from augmented inputs described in the previous section (Eqn-\ref{eqn: ensem_ent}). First, we generate a selective entropy map by taking a weighted combination of no augmentation-based entropy map $H(\hat{y}_t^n)$ and augmentation-based entropy maps $H(\tilde{Y}_t^n)$:
\begin{equation}
    Z_S = Norm( \alpha*H(\hat{y}_t^n) + (1-\alpha)*H(\tilde{Y}_t^n) ).
\end{equation}
where $\alpha$ is a hyper-parameter. Next, we generate a false negative region mask $U_t^n$ from $p(\hat{y}_t^n)$ as follows:
\begin{equation}
U_t^n=\left\{\begin{array}{ll}
0, & \;\;\;p(\hat{y}_t^n) < \lambda_1 \:  \& \:  p(\hat{y}_t^n) > \lambda_2\\
 1 , & \;\;\;\lambda_1 < p(\hat{y}_t^n) < \lambda_2.
\end{array}\right.
\end{equation}
where $\lambda_1$ and $\lambda_2$ are false negative region threshold values. Finally, we suppress the false negatives in $U_t^n$ by performing element-wise $AND$ operation with the selective entropy mask ($Z_t^n$). The enhanced pseudo-label $\mathbf{\bar{{y}}_t^n}$ is obtained using the following formulation:
\begin{equation}
    \mathbf{\bar{y}_t^n} = \hat{y}_t^n | (Z_t^n \& U_t^n) .
\end{equation}
where $|$ denotes the element-wise $OR$ operation. We visualize the intermediate steps for a sample input $x_t^n$ in Fig. \ref{stage1}. It can be seen that the selectively voted entropy map $H_S$ captures more consistent high entropy regions compared to the normal entropy map $H(\tilde{y}_t^n)$. Our enhanced pseudo-label $\mathbf{\bar{{y}}_t^n}$ can also be observed to be a refined version of the original pseudo-label $\hat{y}_t^n$.
To train $\Theta_t^{stage1}$, we define the stage-I training loss $\mathcal{L}_{stage1}$ as:
\begin{equation}
    \mathcal{L}_{stage1} = \mathcal{L}_{seg}(x_t^n, \mathbf{\bar{{y}}_t^n}) + \mathcal{L}_{EEM}.
\end{equation}
\vspace{-1.0 em}

\vspace{-1.0 em}
\subsection{Task-specific Adaptation}
After performing target-specific adaptation, our model is able to generate high-quality pseudo-labels. We use the enhanced pseudo-labels to improve the segmentation performance in the target domain. This is done by utilizing a Student-Teacher model based on semi-supervised learning techniques, as described in Wang et al. (2021). Our approach involves applying both strong and weak augmentations to input images, where the strongly augmented image is used as input to the student model $\Theta_t^{student}$, while the weakly augmented image is used as input to the teacher model $\Theta_t^{teacher}$. The pseudo-labels generated from the teacher model are then used to supervise the student model, and the teacher model is gradually updated by transferring weights from the student model. It is important to note that both the teacher and student models are initialized with the model obtained from Stage I, i.e., $\Theta_t^{stage1}$. Hence, the stage II loss $\mathcal{L}_{stage2}$ is defined as follows:
\begin{equation}
    \mathcal{L}_{stage2} = \mathcal{L}_{seg}(x_t^n, y_t^n) 
\end{equation}
The adapted model weights at Stage II, denoted as $\Theta_t^{stage2}$, correspond to the final state of the Teacher model $\Theta_t^{teacher}$ after the self-training process.


\section{Experiments and Results}

\noindent \textbf{Datasets:} For 2D experiments, we focus on the task of retinal vessel segmentation from fundus images. We make use of the following datasets: CHASE \cite{fraz2012ensemble}, RITE \cite{hu2013automated} and HRF \cite{odstrvcilik2009improvement}. For 3D experiments, we focus on brain tumor segmentation from MRI volumes. We make use of the BraTS 2019 dataset \cite{menze2014multimodal,bakas2017advancing} which consists of four modalities of MRI- FLAIR, T1, T1ce and T2. We study the domain shift problems between these four modalities for volumetric segmentation of brain tumor. This is a multi-class segmentation problem with 4 labels. We randomly split the dataset into 266 for training and 69 for validation. 

\begin{figure}[htbp]
 \vspace{-1.5 em}
	\centering
	\includegraphics[width=1\linewidth]{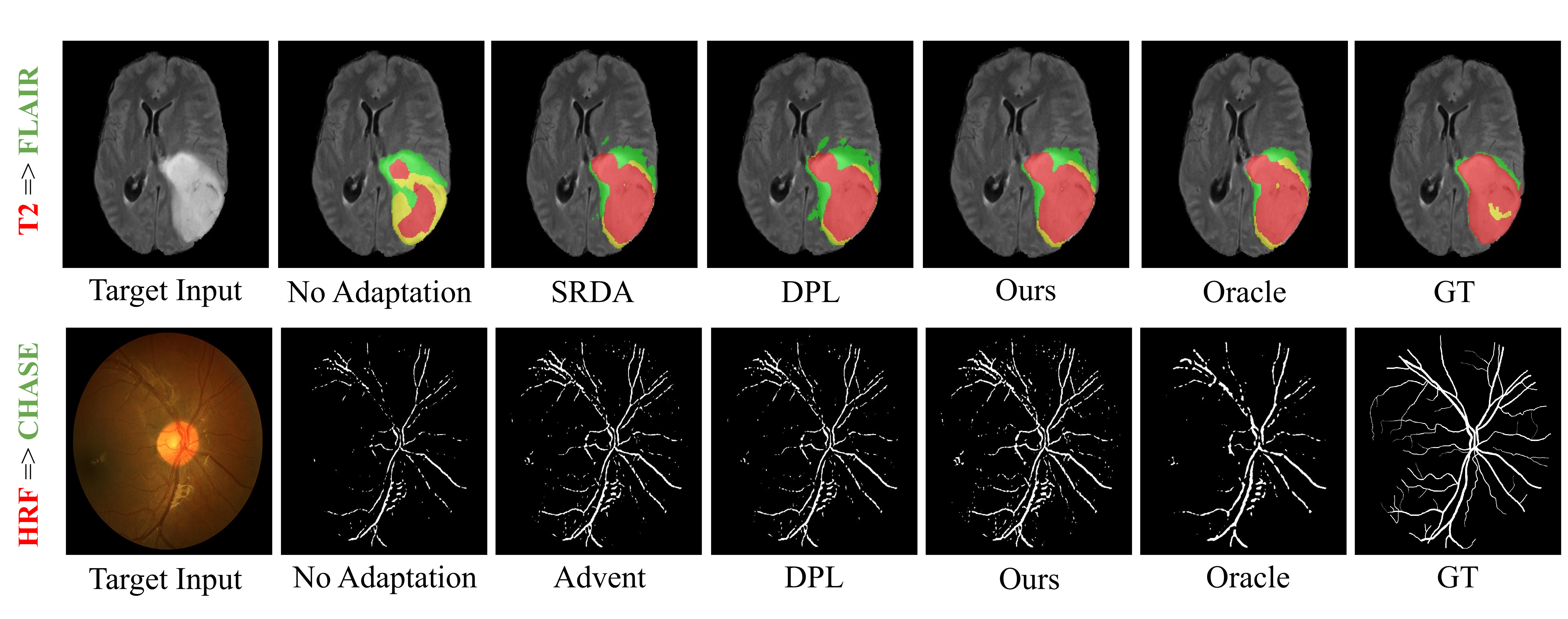}
	\vskip-10pt\caption{Qualitative results for samples from distribution shifts on fundus and BraTS dataset.  Red label: Necrosis, yellow label: Edema and Green label: Edema.}  
	\label{qual}
 \vspace{-1 em}
\end{figure}

\noindent \textbf{Implementation Details: }  In the 2D source experiments, we train a UNet model using the segmentation loss $\mathcal{L}_{seg}(x_s, y_s)$, with an Adam optimizer and a learning rate of 0.001, momentum of 0.9. We also use a cosine annealing learning rate scheduler with a minimum learning rate of 0.0001. For 3D experiments, we use a similar loss but for multi-class segmentation. In TT-SFUDA experiments, we use the Adam optimizer with a learning rate of 0.0001 and momentum of 0.9 for both stages. For Stage I, we use augmentations such as color jitter, grayscale, color contrast, and solarize. For Stage II, we use strong augmentations such as color jitter, grayscale, and horizontal flip, while the weak augmentation consists of only horizontal flip. Our pipeline consists of three hyper-parameters: $\alpha$, $\lambda_1$, and $\lambda_2$, which are set to 0.75, 0.3, and 0.5, respectively. For Stage II, the teacher EMA rate is set to 0.99 for each iteration in the student-teacher network. We train Stage I for one epoch and Stage II for ten epochs. The framework is implemented in PyTorch using NVIDIA Titan X GPU.

\noindent \textbf{Performance Comparison: } To evaluate the effectiveness of our proposed TT-SFUDA method, we compared it against both UDA and SFUDA methods. For UDA, we compare against a pseudo-label self-training \cite{zou2018unsupervised} based approach; BEAL \cite{wang2019boundary} which uses adversarial learning between source and target data with boundary information specifically for fundus images, and AdvEnt \cite{vu2019advent} which ensures entropy consistency between the source and target domains. For SFUDA, we compare against SRDA \cite{bateson2020source} and DPL \cite{chen2021source}, where SRDA uses a task prior while DPL uses an uncertainty-guided denoised pseudo-label based approach. Note that we adopt these methods for 3D by making necessary changes. In Tables \ref{2d} and \ref{3d}, we tabulate our results for domain shifts in 2D fundus images and 3D MRI volumes respectively. We achieve a competitive performance compared to both UDA and SFUDA baselines across almost all shifts.  We present sample qualitative results in Fig. \ref{qual} where we observe a good performance for TT-SFUDA when compared to other UDA and SFUDA methods.

\vspace{-1 em}

\begin{table}[]
\centering
\caption{Results for 2D Domain shifts. Numbers correspond to dice scores. \textbf{\textcolor{red}{Red}} and \textbf{\textcolor{blue}{Blue}} correspond to the first and second best performing methods respectively.}
\resizebox{1\columnwidth}{!}{%
\begin{tabular}{cccccc}
\toprule
Type                    & Method                                                    & CHASE$\rightarrow$HRF & CHASE$\rightarrow$RITE & HRF$\rightarrow$CHASE & HRF$\rightarrow$RITE \\ \hline
Source-Training         & Direct Testing                                            & 52.70                    & 15.45                     & 57.92                    & 41.03                   \\ \hline
\multirow{4}{*}{UDA}    & PL \cite{zou2018unsupervised}                                                    &  53.04 $\pm$ 1.32                       & 34.62  $\pm$  1.50                      & 57.41  $\pm$ 2.01                      & 41.98 $\pm$   1.67                    \\
& BEAL  \cite{wang2019boundary}                                                  & \textbf{\textcolor{red}{60.54 $\pm$ 2.21} }                      & 50.24  $\pm$ 2.86                        & 59.21 $\pm$  3.10                      & 53.43 $\pm$  2.51                     \\
                        & AdvEnt   \cite{vu2019advent}                                                     &    58.23 $\pm$ 2.30                    & 49.91 $\pm$ 2.16                          & 58.50 $\pm$ 1.89                       & 52.49 $\pm$ 2.11                       \\ \hline
\multirow{3}{*}{SFUDA} & SRDA  \cite{bateson2020source}                                                      &  55.76   $\pm$  1.58                   & 50.43 $\pm$  1.40                       & 60.81  $\pm$    1.31                    & \textbf{\textcolor{blue}{56.92   $\pm$ 1.99}}                    \\
                        & DPL \cite{chen2021source}                                                     & 56.32  $\pm$ 1.23                       & \textbf{\textcolor{blue}{51.27   $\pm$ 1.62}}                      & \textbf{\textcolor{blue}{61.29  $\pm$ 1.16 }}                      & 55.89  $\pm$ 1.23                     \\
                        & \textbf{TT-SFUDA} & \textbf{\textcolor{blue}{58.25 $\pm$ 1.06}}                         & \textbf{\textcolor{red}{52.63   $\pm$  1.36}}                      & \textbf{\textcolor{red}{64.95  $\pm$  0.89}}                     & \textbf{\textcolor{red}{58.37    $\pm$ 1.03}}                    \\ \hline
Target-Training         & Oracle                                                    & 67.97                    & 73.70                     & 66.92                    & 73.70                   \\ \bottomrule
\end{tabular}
}
\label{2d}
\end{table}

\vspace{-2.5 em}

\begin{table}[]
\centering
\caption{Results for 3D Domain shifts. Numbers correspond to dice scores reported in the following order:  WT/TC/ET. WT = Whole Tumor, TC = Tumor Core, ET = Enhancing Tumor. \textbf{\textcolor{red}{Red}} and \textbf{\textcolor{blue}{Blue}} correspond to the first and second best performing methods respectively.}
\resizebox{1.0\columnwidth}{!}{%
\begin{tabular}{ccccc}
\toprule
Type                    & Method         & T2 --\textgreater FLAIR & T1 --\textgreater T1ce & FLAIR --\textgreater T2 \\ \hline
Source-Training         & Direct Testing & 52.60/13.70/37.25       & 30.96/12.33/21.08      & 43.11/12.79/34.84       \\ \hline
               & AdvEnt \cite{vu2019advent}            & 54.51/14.07/38.50                       &  32.40/13.38/21.99                      &    44.01/12.83/35.27                     \\
  \multirow{3}{*}{SFUDA}                      & PL  \cite{zou2018unsupervised}           &      52.95/13.80/37.38                   & 31.04/12.46/21.17                       &   43.23/12.67/34.79                      \\ 
 & SRDA  \cite{bateson2020source}          &   57.23/14.02/38.91                      &    \textbf{\textcolor{blue}{33.98}}/13.76/\textbf{\textcolor{red}{22.67}}                    &     43.17/12.52/\textbf{\textcolor{blue}{35.27}}                   \\
                        & DPL \cite{chen2021source}           &       \textbf{\textcolor{blue}{58.32}}/\textbf{\textcolor{blue}{14.10}}/\textbf{\textcolor{red}{41.13}}                 &    33.62/\textbf{\textcolor{red}{14.08}}/\textbf{\textcolor{blue}{22.35}}                    &   \textbf{\textcolor{red}{44.93}}/\textbf{\textcolor{red}{13.04}}/\textbf{\textcolor{red}{35.86}}                      \\
                        & \textbf{TT-SFUDA}      &    \textbf{\textcolor{red}{59.06}}/\textbf{\textcolor{red}{14.16}}/\textbf{\textcolor{blue}{40.67}}                     &   \textbf{\textcolor{red}{34.21}}/\textbf{\textcolor{blue}{13.95}}/22.18                     &       \textbf{\textcolor{blue}{44.46}}/\textbf{\textcolor{blue}{12.83}}/34.83                  \\ \hline
Target-Training         & Oracle         & 75.50/26.57/57.27       & 53.28/13.63/35.26      & 83.95/22.40/51.51       \\ \bottomrule
\end{tabular}
}
\label{3d}
\vspace{-1 em}
\end{table}

\vspace{-1.5 em}

\section{Discussion: }

\noindent \textbf{Stage I vs Stage II: } The results of various stage-wise experiments on 2D domain shifts are presented in Table \ref{abl:stage}, where each row corresponds to a different sequence of stage-wise adaptation performances. Our findings show that the best performance is achieved when task-specific adaptation is performed after target-specific adaptation, which is evident in the Stage I $\rightarrow$ Stage II experiment. On the other hand, when we first learn task-specific representation and then perform entropy minimization for target-specific adaptation, the model becomes overfitted to the noise generated from the pseudo-labels, leading to a significant drop in performance. Furthermore, we observe that the drop in performance in Stage II $\rightarrow$ Stage I experiment is directly proportional to the difference in the domain gap. Overall, our results highlight the importance of the order in which the stages of adaptation are performed to achieve optimal performance.
\vspace{-1.0 em}
\begin{table}[htbp]
\centering
\caption{Analysis on sequence of stage-wise experiments for 2D Domain shifts.}
\resizebox{1\columnwidth}{!}{%
\begin{tabular}{cccccc}
\hline
Type                    & Method                                                    & CHASE$\rightarrow$HRF & CHASE$\rightarrow$RITE & HRF$\rightarrow$CHASE & HRF$\rightarrow$RITE \\ \hline
Source-Training         & Direct Testing                                            & 52.70                    & 15.45                     & 57.92                    & 41.03                   \\ \hline
\multirow{3}{*}{TT-SFUDA}  & Stage II $\rightarrow$ Stage I & 55.20  $\pm$ 1.45                       & 18.28  $\pm$  2.26                       & 63.67 $\pm$ 1.70                       & 49.02 $\pm$ 2.08                       \\
                & Stage I $\rightarrow$ Stage II &       58.25 $\pm$ 1.06                        & 52.63   $\pm$  1.36                     & 64.95  $\pm$  0.89                   & 58.37    $\pm$ 1.03                     \\
                        \hline
Target-Training         & Oracle                                                    & 67.97                    & 73.70                     & 66.92                    & 73.70                   \\ \hline
\end{tabular}
}
\label{abl:stage}
\end{table}

\vspace{-1.0 em}

\noindent \textbf{Ablation study: } Table \ref{abl:mod} shows the impact of different loss functions used in our pipeline. In Stage I, we use ensemble entropy minimization (EEM) loss and Selective Voting (SV) loss, while in Stage II, we use student-teacher (ST) loss.  We can observe that using EEM loss helps the model to adapt towards the target domain by minimizing the entropy. However, by incorporating selective voting and training with enhanced pseudo-labels, we can observe a further improvement in performance. This suggests that selective voting is a useful technique for improving the quality of the pseudo-labels along with EEM. Moreover, if we apply only the student-teacher loss in Stage II, we can see a huge boost in performance in the domain where there is a large domain gap. It is worth noting that the student-teacher frameworks play a crucial role in performance boosting when there is a significant domain gap. Finally, by leveraging all the proposed losses in both stages of training results in the best performance. 

\vspace{-0.5 em}

\begin{table}[htbp]
\centering
\caption{Ablation study on different loss functions i.e ensemble entropy minimization (EEM), selective voting (SV) and student-teacher (ST) loss for 2D Domain shifts.}
\resizebox{1\columnwidth}{!}{%
\begin{tabular}{cccccccc}
\hline
Type                    & EEM & SV &   ST                                                  & CHASE$\rightarrow$HRF & CHASE$\rightarrow$RITE & HRF$\rightarrow$CHASE & HRF$\rightarrow$RITE \\ \hline
Source-Training         & \xmark &  \xmark &  \xmark                                            & 52.70                    & 15.45                     & 57.92                    & 41.03                   \\ \hline
\multirow{3}{*}{TT-SFUDA} &  \cmark & \xmark &  \xmark          &  55.23   $\pm$ 0.65                    &    25.35 $\pm$ 1.01                        &    59.28   $\pm$ 0.85                      & 47.13  $\pm$   1.07                   \\
                        &  \cmark & \cmark &  \xmark                 &  56.17   $\pm$ 1.08                    & 27.19 $\pm$ 1.35                        & 60.18   $\pm$ 1.10                      & 48.07  $\pm$   1.32               \\
                        &  \xmark & \xmark &  \cmark   & 54.62 $\pm$ 1.52                       & 39.28  $\pm$  2.13                      & 58.40 $\pm$  1.17                        & 52.97  $\pm$    1.63                       \\
                &  \cmark & \cmark &  \cmark   &   58.25 $\pm$ 1.06                         & 52.63   $\pm$  1.36                     & 64.95  $\pm$  0.89                    &  58.37    $\pm$ 1.03                    \\
                        \hline
Target-Training        & & &                                                     & 67.97                    & 73.70                     & 66.92                    & 73.70                   \\ \hline
\end{tabular}
}
\label{abl:mod}
\end{table}

 \vspace{-2.5 em}

\section{Conclusion}
In summary, we have introduced a systematic approach for source-free unsupervised domain adaptive image segmentation called TT-SFUDA. Our method involves a two-stage process where we first perform target-specific adaptation by minimizing high entropy regions using an ensemble entropy minimization loss and a selective voting strategy. This is followed by task-specific adaptation where we use a student-teacher framework to effectively learn segmentation on the target domain by leveraging the enhanced pseudo-labels. Our experimental analysis shows that our method outperforms existing UDA and SFUDA methods on multiple domain shifts on both 2D and 3D datasets, demonstrating the effectiveness of our approach.

\bibliographystyle{splncs04}
\bibliography{sfuda}

\begin{thebibliography}{10}
\providecommand{\url}[1]{\texttt{#1}}
\providecommand{\urlprefix}{URL }
\providecommand{\doi}[1]{https://doi.org/#1}

\bibitem{bakas2017advancing}
Bakas, S., Akbari, H., Sotiras, A., Bilello, M., Rozycki, M., Kirby, J.S.,
  Freymann, J.B., Farahani, K., Davatzikos, C.: Advancing the cancer genome
  atlas glioma mri collections with expert segmentation labels and radiomic
  features. Scientific data  \textbf{4}(1),  1--13 (2017)

\bibitem{bateson2020source}
Bateson, M., Kervadec, H., Dolz, J., Lombaert, H., Ben~Ayed, I.: Source-relaxed
  domain adaptation for image segmentation. In: International Conference on
  Medical Image Computing and Computer-Assisted Intervention. pp. 490--499.
  Springer (2020)

\bibitem{chen2021source}
Chen, C., Liu, Q., Jin, Y., Dou, Q., Heng, P.A.: Source-free domain adaptive
  fundus image segmentation with denoised pseudo-labeling. In: International
  Conference on Medical Image Computing and Computer-Assisted Intervention. pp.
  225--235. Springer (2021)

\bibitem{chen2021transunet}
Chen, J., Lu, Y., Yu, Q., Luo, X., Adeli, E., Wang, Y., Lu, L., Yuille, A.L.,
  Zhou, Y.: Transunet: Transformers make strong encoders for medical image
  segmentation. arXiv preprint arXiv:2102.04306  (2021)

\bibitem{cciccek20163d}
{\c{C}}i{\c{c}}ek, {\"O}., Abdulkadir, A., Lienkamp, S.S., Brox, T.,
  Ronneberger, O.: 3d u-net: learning dense volumetric segmentation from sparse
  annotation. In: International conference on medical image computing and
  computer-assisted intervention. pp. 424--432. Springer (2016)

\bibitem{fraz2012ensemble}
Fraz, M.M., Remagnino, P., Hoppe, A., Uyyanonvara, B., Rudnicka, A.R., Owen,
  C.G., Barman, S.A.: An ensemble classification-based approach applied to
  retinal blood vessel segmentation. IEEE Transactions on Biomedical
  Engineering  \textbf{59}(9),  2538--2548 (2012)

\bibitem{hatamizadeh2022unetr}
Hatamizadeh, A., Tang, Y., Nath, V., Yang, D., Myronenko, A., Landman, B.,
  Roth, H.R., Xu, D.: Unetr: Transformers for 3d medical image segmentation.
  In: Proceedings of the IEEE/CVF Winter Conference on Applications of Computer
  Vision. pp. 574--584 (2022)

\bibitem{hong2022source}
Hong, J., Zhang, Y.D., Chen, W.: Source-free unsupervised domain adaptation for
  cross-modality abdominal multi-organ segmentation. Knowledge-Based Systems
  \textbf{250},  109155 (2022)

\bibitem{hu2013automated}
Hu, Q., Abr{\`a}moff, M.D., Garvin, M.K.: Automated separation of binary
  overlapping trees in low-contrast color retinal images. In: International
  conference on medical image computing and computer-assisted intervention. pp.
  436--443. Springer (2013)

\bibitem{islam2018ischemic}
Islam, M., Vaidyanathan, N.R., Jose, V., Ren, H.: Ischemic stroke lesion
  segmentation using adversarial learning. In: International MICCAI Brainlesion
  Workshop. pp. 292--300. Springer (2018)

\bibitem{javanmardi2018domain}
Javanmardi, M., Tasdizen, T.: Domain adaptation for biomedical image
  segmentation using adversarial training. In: 2018 IEEE 15th International
  Symposium on Biomedical Imaging (ISBI 2018). pp. 554--558. IEEE (2018)

\bibitem{liu2023memory}
Liu, X., Xing, F., El~Fakhri, G., Woo, J.: Memory consistent unsupervised
  off-the-shelf model adaptation for source-relaxed medical image segmentation.
  Medical Image Analysis  \textbf{83},  102641 (2023)

\bibitem{menze2014multimodal}
Menze, B.H., Jakab, A., Bauer, S., Kalpathy-Cramer, J., Farahani, K., Kirby,
  J., Burren, Y., Porz, N., Slotboom, J., Wiest, R., et~al.: The multimodal
  brain tumor image segmentation benchmark (brats). IEEE transactions on
  medical imaging  \textbf{34}(10),  1993--2024 (2014)

\bibitem{milletari2016v}
Milletari, F., Navab, N., Ahmadi, S.A.: V-net: Fully convolutional neural
  networks for volumetric medical image segmentation. In: 2016 fourth
  international conference on 3D vision (3DV). pp. 565--571. IEEE (2016)

\bibitem{odstrvcilik2009improvement}
Odstr{\v{c}}il{\'\i}k, J., Jan, J., Gaz{\'a}rek, J., Kol{\'a}{\v{r}}, R.:
  Improvement of vessel segmentation by matched filtering in colour retinal
  images. In: World Congress on Medical Physics and Biomedical Engineering,
  September 7-12, 2009, Munich, Germany. pp. 327--330. Springer (2009)

\bibitem{panfilov2019improving}
Panfilov, E., Tiulpin, A., Klein, S., Nieminen, M.T., Saarakkala, S.: Improving
  robustness of deep learning based knee mri segmentation: Mixup and
  adversarial domain adaptation. In: Proceedings of the IEEE/CVF International
  Conference on Computer Vision Workshops. pp.~0--0 (2019)

\bibitem{perone2019unsupervised}
Perone, C.S., Ballester, P., Barros, R.C., Cohen-Adad, J.: Unsupervised domain
  adaptation for medical imaging segmentation with self-ensembling. NeuroImage
  \textbf{194},  1--11 (2019)

\bibitem{ronneberger2015u}
Ronneberger, O., Fischer, P., Brox, T.: U-net: Convolutional networks for
  biomedical image segmentation. In: International Conference on Medical image
  computing and computer-assisted intervention. pp. 234--241. Springer (2015)

\bibitem{jose2021medical}
Valanarasu, J.M.J., Oza, P., Hacihaliloglu, I., Patel, V.M.: Medical
  transformer: Gated axial-attention for medical image segmentation. In:
  Medical Image Computing and Computer Assisted Intervention -- MICCAI 2021.
  pp. 36--46. Springer International Publishing, Cham (2021)

\bibitem{valanarasu2020kiu}
Valanarasu, J.M.J., Sindagi, V.A., Hacihaliloglu, I., Patel, V.M.: Kiu-net:
  Towards accurate segmentation of biomedical images using over-complete
  representations. In: Medical Image Computing and Computer Assisted
  Intervention--MICCAI 2020: 23rd International Conference, Lima, Peru, October
  4--8, 2020, Proceedings, Part IV 23. pp. 363--373. Springer (2020)

\bibitem{vashist2017point}
Vashist, S.K.: Point-of-care diagnostics: Recent advances and trends.
  Biosensors  \textbf{7}(4), ~62 (2017)

\bibitem{vu2019advent}
Vu, T.H., Jain, H., Bucher, M., Cord, M., P{\'e}rez, P.: Advent: Adversarial
  entropy minimization for domain adaptation in semantic segmentation. In:
  Proceedings of the IEEE/CVF Conference on Computer Vision and Pattern
  Recognition. pp. 2517--2526 (2019)

\bibitem{wang2021knowledge}
Wang, L., Yoon, K.J.: Knowledge distillation and student-teacher learning for
  visual intelligence: A review and new outlooks. IEEE Transactions on Pattern
  Analysis and Machine Intelligence  (2021)

\bibitem{wang2019boundary}
Wang, S., Yu, L., Li, K., Yang, X., Fu, C.W., Heng, P.A.: Boundary and
  entropy-driven adversarial learning for fundus image segmentation. In:
  Medical Image Computing and Computer Assisted Intervention--MICCAI 2019: 22nd
  International Conference, Shenzhen, China, October 13--17, 2019, Proceedings,
  Part I 22. pp. 102--110. Springer (2019)

\bibitem{yang2019unsupervised}
Yang, J., Dvornek, N.C., Zhang, F., Chapiro, J., Lin, M., Duncan, J.S.:
  Unsupervised domain adaptation via disentangled representations: Application
  to cross-modality liver segmentation. In: International Conference on Medical
  Image Computing and Computer-Assisted Intervention. pp. 255--263. Springer
  (2019)

\bibitem{zhang2019unsupervised}
Zhang, J., Liu, M., Pan, Y., Shen, D.: Unsupervised conditional consensus
  adversarial network for brain disease identification with structural mri. In:
  International Workshop on Machine Learning in Medical Imaging. pp. 391--399.
  Springer (2019)

\bibitem{zou2018unsupervised}
Zou, Y., Yu, Z., Kumar, B., Wang, J.: Unsupervised domain adaptation for
  semantic segmentation via class-balanced self-training. In: Proceedings of
  the European conference on computer vision (ECCV). pp. 289--305 (2018)

\end{thebibliography}

\end{document}